\documentclass[conference]{IEEEtran}
\IEEEoverridecommandlockouts

\usepackage{fix-cm}
\usepackage{amsmath}   
\usepackage{booktabs}  
\usepackage{graphicx}  
\usepackage{multirow}

\usepackage{cite}
\usepackage{amsmath,amssymb,amsfonts}
\usepackage{algorithmic}
\usepackage{graphicx}
\usepackage{textcomp}
\usepackage{xcolor}
\def\BibTeX{{\rm B\kern-.05em{\sc i\kern-.025em b}\kern-.08em
    T\kern-.1667em\lower.7ex\hbox{E}\kern-.125emX}}
\begin{document}

\title{Meta-Ensemble Learning with Diverse Data Splits for Improved Respiratory Sound Classification\\
\thanks{\hspace{-1em}$^{\dagger}$ Corresponding author. This work was partially supported by the National Research Foundation of Korea(NRF) grant funded by the Korea government(MSIT) (Grant no. RS-2025-16066662), and by the Regional Innovation System \& Education(RISE) program through the Jeonbuk RISE Center, funded by the Ministry of Education(MOE) and the Jeonbuk State, Republic of Korea (2026-RISE-13-WKU).}
\thanks{\hspace{-1em}$^{1}$Department of Electronic Engineering, Wonkwang University, Republic of Korea $^{2}$RSC LAB, MODULABS, Republic of Korea $^{3}$AICU Global Inc., Republic of Korea $^{4}$Seoul National University Bundang Hospital, Republic of Korea.}
}

\author{June-Woo Kim$^{1}$ Miika Toikkanen$^{2}$ Heejoon Koo$^{2}$ Yoon Tae Kim$^{2}$ Doyoung Kwon$^{3}$ Kyunghoon Kim$^{4}$$^{\dagger}$
\\ kaen2891@wku.ac.kr \quad journey237@snu.ac.kr
}

\maketitle

\begin{abstract}
Training reliable respiratory sound classification models remains challenging due to the limited size and subject diversity of datasets. Ensemble methods can improve robustness, but when base models are trained on identical data, models tend to overfit and produce highly correlated predictions, thereby reducing the effectiveness of ensembling. In this work, we investigate a meta-ensemble learning methodology that enhances prediction diversity by training base models on diverse data splits and combining their outputs through a trained meta-model. Specifically, we train base models on the ICBHI dataset using two data split settings: \emph{fixed} 80--20\% split and \emph{five-fold cross-validation} split, under two data granularity settings: \emph{patient}- and \emph{sample}-level. The resulting diversity in base model predictions enables the meta-model to better generalize. Our approach achieves new state-of-the-art performance on the ICBHI benchmark, reaching a Score of 66.49\% and showing improved generalization on two out-of-distribution datasets, indicating its potential applicability to real-world clinical data.
\end{abstract}


\section{Introduction}
Respiratory sound classification (RSC) is important for diagnosing conditions such as asthma, chronic obstructive pulmonary disease, and pneumonia~\cite{kim2021respiratory}. However, the development of reliable RSC models remains difficult due to the limited availability of annotated data, constrained by privacy issues, high collection costs, and the need for expert labeling~\cite{bae23b_interspeech, xia2022exploring}. Data scarcity poses a major challenge for deep learning, as limited training sets often lead to overfitting and poor generalization, especially in clinical applications where robustness across diverse patients is critical~\cite{kim2024stethoscope, kim2025adaptive, koo2026empowering}. 

Ensemble methods have been explored to improve generalization on limited datasets such as the ICBHI benchmark~\cite{rocha2018alpha, nguyen2022lung, nadkarni2024afen, toikkanen25_interspeech}, but training all base models on the same distribution yields highly correlated errors and reduces ensemble effectiveness~\cite{ortega2022diversity, wood2023unified}. Moreover, fixed fusion schemes like logit averaging cannot account for the varying reliability of base models~\cite{ju2018relative}, motivating the use of meta-models that learn optimal combinations from training or held-out splits~\cite{tsymbal2005diversity}.

Building on these insights, we propose a meta-ensemble learning approach that leverages data variation to improve RSC. Specifically, our strategy begins by dividing the training data into two subsets: 80\% for base model training and 20\% reserved for meta-model training. We then consider two ways of employing the 80\% portion: $(i)$ a \emph{fixed split}, where all base models are trained on the same 80\%, and $(ii)$ a \emph{five-fold cross-validation}, where the 80\% subset is further partitioned to increase variation across base models. 
Within each setting, we apply two levels of granularity: \emph{patient-level} (no subject overlap) and \emph{sample-level} (random sampling regardless of patient). This yields four distinct partitioning strategies for training base models, and the resulting diversity in base models outputs enables the meta-model to learn a more effective combination strategy that improves generalization.

\begin{figure}[t!]
    \centering
    \includegraphics[width=1.0\linewidth]{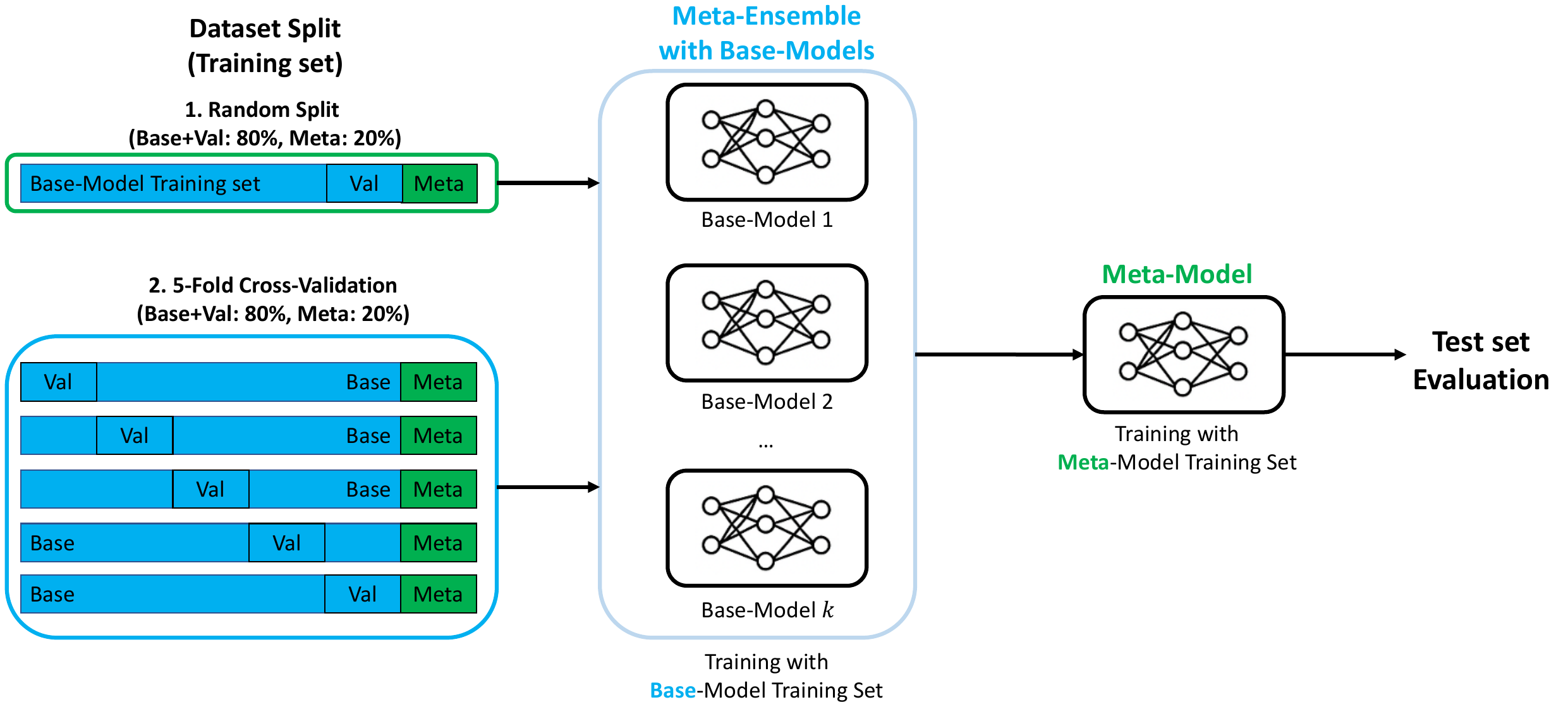}
    \caption{Overview of the proposed meta-ensemble framework, where diverse base models are trained under different splits and aggregated by a meta-model for final prediction.}
    \label{fig:architecture}
\end{figure}

We validate our strategy by constructing ensembles of base-models built upon the prior state-of-the-art BTS model~\cite{kim24f_interspeech} on the ICBHI dataset, and assessing their performance through various meta-model architectures.
Our results demonstrate that incorporating data diversity at the base-model level leads to improved generalization, with the meta-model serving as a critical component for enhancing ensemble performance. Our best configuration achieves an ICBHI score of 66.49\%, and our method exhibits generalization on two out-of-distribution (OOD) datasets: the SPRSound~\cite{zhang2022sprsound} benchmark and the SNUBH~\cite{kim2025adaptive} in-house clinical dataset.

The main contributions of this paper are summarized as follows:

\begin{itemize}
    \item We demonstrate a meta-ensemble learning method that explicitly leverages data diversity by training base classifiers on distinct data splits, including results on both patient-level and sample-level partitions.
    \item We systematically evaluate various meta-model architectures, such as feedforward networks and Transformer-based~\cite{vaswani2017attention} meta-models, to effectively integrate the predictions of heterogeneous base models.
    \item We demonstrate that incorporating data diversity at the base model level substantially improves generalization, outperforming standard ensemble baselines on the ICBHI 2017 Challenge dataset.
    \item We show that the proposed approach achieves robust performance on a held-out validation set collected from real-world clinical settings, underscoring its practical applicability to RSC tasks.
\end{itemize}
\section{Related Works}

\subsection{Respiratory Sound Classification}
The respiratory sound classification task has attracted growing attention for its potential to support early diagnosis of pulmonary diseases. Due to the difficulty and cost of collecting labeled respiratory data, most public datasets, such as the ICBHI 2017 Challenge dataset~\cite{rocha2018alpha} remain limited in both size and patient diversity. To overcome this, prior work has focused on designing model architectures and leveraging improved learning techniques. 
Early models typically adopted convolutional architectures such as ResNet~\cite{he2016deep}, EfficientNet~\cite{tan2019efficientnet}, and CNN6~\cite{kong2020panns}. Later studies demonstrated the effectiveness of pretrained vision and audio models, particularly those trained on large-scale datasets like ImageNet~\cite{deng2009imagenet} and AudioSet~\cite{audioset}. The Audio Spectrogram Transformer (AST)~\cite{gong21b_interspeech, bae23b_interspeech} extended this approach using self-attention to model temporal patterns in spectrograms. Techniques such as SpecAugment~\cite{park2019specaugment} have been used to augment spectrograms, while more recent work explores patch-level~\cite{bae23b_interspeech} and representation-level augmentation~\cite{10782363} and stethoscope bias mitigation~\cite{kim2024stethoscope}. Other studies have proposed leveraging audio-text alignment~\cite{kim24f_interspeech}, adversarial domain adaptation~\cite{kim2023adversarial}, or large-scale pretraining~\cite{wu2023large, niizumi2024masked, zhang2024towards} to further improve generalization.

\subsection{Ensemble Learning}
Ensemble learning methods such as bagging~\cite{breiman1996bagging}, boosting~\cite{freund1997decision}, and stacking~\cite{wolpert1992stacked} have long been used to improve model robustness and generalization.
Ensembling has emerged as a practical method to improve performance at the cost of compute~\cite{li2023towards} and to increase robustness in the face of small, noisy datasets~\cite{dietterich2000ensemble}. 
Unlike the fixed predictor combination strategies, such as averaging of logits, a meta-model can learn from base models' confidence and complementarity, often improving generalization. This meta-ensemble learning, also known as stacked generalization~\cite{wolpert1992stacked}, involves training a second-level model to combine the predictions of multiple base learners. However, stacking is most effective when base models are diverse. If they make identical or highly correlated predictions, 
the meta-model offers limited advantage over simpler combination strategies.

Our study explicitly addresses the challenge of base model homogeneity in ensemble learning by introducing data diversity through distinct training splits. By combining this strategy with a meta-ensemble framework, we enhance base model complementarity and enable the meta-model to more exploit the strengths and compensate for the weaknesses of individual predictors, ultimately improving generalization performance.
\section{Preliminaries}

\subsection{Dataset Description}
We used the ICBHI~\cite{rocha2018alpha} dataset for model training, while SPRSound~\cite{zhang2022sprsound} and SNUBH~\cite{kim2025adaptive} served as OOD test sets to evaluate generalization.
\noindent \textbf{ICBHI} respiratory sound dataset~\cite{rocha2018alpha} is a well-established benchmark in the field of RSC. The dataset comprises approximately 5.5 hours of respiratory recordings, encompassing a total of 6,898 breathing cycles. For evaluation consistency, the dataset is officially partitioned at the breathing cycle level into non-overlapping training (60\%) and testing (40\%) sets, yielding 4,142 cycles for training and 2,756 for testing. The data are annotated into four categories: \textit{normal}, \textit{crackle}, \textit{wheeze}, and \textit{both} (crackle and wheeze). Following previous work~\cite{kim24f_interspeech}, we divided the age variable into adult (older than 18 years) and pediatric (18 years or younger) groups. Additional metadata, such as patient gender, recording location, and recording device type (electronic stethoscope), were preserved according to the original dataset annotations.

\noindent \textbf{SPRSound} is a widely used RSC dataset consisting of Chinese pediatric patients. While it includes 8,085 segments (approximately 11 hours in total) spanning seven respiratory sound classes, we align it with the 4-class taxonomy of ICBHI-trained models by merging \emph{coarse crackle} and \emph{fine crackle} into a single \emph{crackle} class, as well as \emph{stridor} and \emph{rhonchi} into the \emph{wheeze} category. We used the official test set for OOD evaluation.

\noindent \textbf{SNUBH} dataset is our in-house held-out clinical dataset focused on early-stage asthma diagnosis in Korean pediatric patients. Unlike the ICBHI, annotations were provided at the recording level rather than at the cycle level. This dataset contains 2,134 respiratory sound instances (4.2 hours in total), of which 418 test samples were used for evaluation, following the same protocol as in~\cite{kim2025adaptive}. Classification was performed between the \emph{wheeze} and \emph{other} classes, and evaluation metrics were computed in the same manner for ICBHI.

\subsection{Training Details}
Respiratory cycles were extracted from the raw waveforms and standardized to 8-second segments, following the procedures outlined in prior studies~\cite{bae23b_interspeech, kim2023adversarial, 10782363, kim24f_interspeech, kim2025adaptive, kim2025tri}. We followed the same original experimental configurations as described in the BTS~\cite{kim24f_interspeech} work. Therefore, all respiratory sound samples were resampled to 48 kHz to align with the input requirements of the BTS model. For the base model training, we used the Adam optimizer with a learning rate of 5e--5, cosine learning rate scheduling, and a batch size of 8 until 50 epochs. The meta-model was fine-tuned for 10 epochs using the same training configurations as applied to the base model.

\subsection{Evaluation Metrics}
Performance evaluation was conducted using~\textit{Sensitivity} ($S_e$), \textit{Specificity} ($S_p$), and their arithmetic mean, referred to as the ICBHI \textit{Score}, following the definitions provided in~\cite{rocha2018alpha}. Sensitivity and specificity respectively measure the correctly identified abnormal and normal respiratory cases. We report both the mean and standard deviation of $S_p$, $S_e$, and Score across five independent runs with random seeds (1--5). We further report the \textit{Relative Rate of Change} (RRC, \%), representing the performance gain relative to the base models' mean performance. 
For mean-ensemble models, results are deterministic given the base models and are therefore reported from a single run.
\section{Methodology}

We apply a Meta-Ensemble learning strategy that improves generalization by producing diverse base model predictions and using a trainable combination method that benefits more from diversity. 

\subsection{Baseline: BTS}
We use BTS~\cite{kim24f_interspeech} as the base model architecture. BTS leverages the pretrained LAION-CLAP model~\cite{wu2023large} to extract embeddings from both respiratory sound and associated metadata, which includes patients' age group, sex, recording locations, and recording devices. These embeddings are fused and passed through a shallow classifier to predict the lung sound class labels.

\subsection{Data Strategy}
Ensemble methods generally combine models trained on the same data, which limits their ability to generalize. In contrast, we aim to improve ensemble performance by increasing diversity among base models by using different data partitions.



\subsubsection{Split Method}
We first divide the original training set into two subsets: 80\% for base models training and 20\% for meta-model training, with no overlap between them. 
Based on how the 80\% portion is employed, we examine two strategies:

\begin{itemize}
    \item Fixed split: All five base models are trained independently on the same 80\% subset, with variation arising only from random initialization by running seeds.
    \item Five-fold cross-validation: To induce greater diversity, the 80\% subset is further partitioned into five folds. Each base model is trained on a unique combination of four folds and validated on the remaining fold. Although each base model is trained on a slightly smaller portion of the data compared to the fixed split, the variation introduced by different train--validation configurations yields greater diversity among base models. As before, the 20\% subset for meta-model training remains fixed and is unseen by any base model. 
\end{itemize}



\subsubsection{Split Granularity}
We additionally control the granularity of splitting at two levels:
\begin{itemize}
    \item Patient-Level (P-Level): In this configuration, all samples from a given patient are allocated exclusively to either the base model training set or the meta-model training set, \emph{ensuring no subject overlap}.
This ensures that the model cannot rely on patient-specific cues, thus evaluating its ability to generalize to completely unseen individuals, which is a critical capability for clinical deployment.
    \item Sample-Level (S-Level): Samples are randomly split without considering patient identity, which allows different samples from the same patient to appear in both the base- and meta-model sets. This setting captures intra-patient variability but risks information leakage, potentially inflating performance estimates. Comparing P- and S-levels thus denotes the impact of patient overlap on generalization.
\end{itemize}

This ensures that the model cannot rely on patient-specific cues, thus evaluating its ability to generalize to completely unseen individuals, which is a critical capability for clinical deployment.

\begin{table*}[!t]
    \centering
    \caption{The main results present model comparisons on the ICBHI dataset using the official 60--40\% train-test split. Pretraining Data column, IN, AS, and LA correspond to ImageNet~\cite{deng2009imagenet}, AudioSet~\cite{audioset}, and LAION-Audio-630K~\cite{wu2023large}, respectively. An asterisk $*$ indicates the previously reported state-of-the-art ICBHI Score. The \textbf{Best} and {\underline{second best}} results are highlighted in bold and underlined, respectively.
    }
    
    \label{table1}
    \renewcommand{\arraystretch}{1}
    \addtolength{\tabcolsep}{8pt}
    \resizebox{\linewidth}{!}{
    \begin{tabular}{llll|lll}
    \toprule
    Method & Backbone & Pretraining Data & Venue & $S_p$\,(\%) & $S_e$\,(\%) & \textbf{Score}\,(\%) \\
    \hline \midrule


    

    Chang \textit{et al.} \cite{chang22h_interspeech} & CNN8-dilated & - & \textit{INTERSPEECH`22} & 69.92 & 35.85 & 52.89 \\
    
    

    
   
    
    
    Bae \textit{et al.} \cite{bae23b_interspeech}\, (Patch-Mix CL) & AST & IN\,+\,AS & \textit{INTERSPEECH`23} & $\text{81.66}$ & $\text{{43.07}}$ & $\text{62.37}$ \\
    
    
    Kim \textit{et al.} \cite{kim2024stethoscope}\, (SG-SCL) & AST & IN\,+\,AS & \textit{ICASSP`24} & $\text{{79.87}}$ & $\text{{43.55}}$ & $\text{{61.71}}$ \\



    Daisuke \textit{et al.} \cite{niizumi2024masked}\, (M2D-X/0.7) & M2D ViT & AS & \textit{TASLP`24} & $\text{81.51}$ & $\text{{45.08}}$ & $\text{63.29}$ \\

    Kim \textit{et al.} \cite{kim24f_interspeech}\, (Audio-CLAP) & CLAP & LA & \textit{INTERSPEECH`24} & $\text{80.85}$ & $\text{44.67}$ & $\text{62.56}$ \\

    Kim \textit{et al.} \cite{kim24f_interspeech}\, (BTS) & CLAP & LA & \textit{INTERSPEECH`24} & $\text{81.40}$ & $\underline{\text{{45.67}}}$ & $\text{63.54}$  \\

    Toikkanen \textit{et al.} \cite{toikkanen25_interspeech}\, (BTS-d) & CLAP & LA & \textit{INTERSPEECH`25} & $\text{82.89}$ & ${\textbf{{45.90}}}$ & $\text{64.39}$ \\

    Toikkanen \textit{et al.} \cite{toikkanen25_interspeech}\, (BTS++) & CLAP & LA & \textit{INTERSPEECH`25} & $\text{89.49}$ & ${\text{{41.89}}}$ & $\text{65.69}^\textbf{*}$ \\


    
    \midrule

    \textbf{Mean-Ensemble (Fixed Split + S-level) [ours]} & \text{CLAP} & LA & - & $\textbf{\text{89.87}}$ & 
    $\text{42.82}$ & 
    $\underline{\text{66.34}}$ \\

    \textbf{Mean-Ensemble (5-Fold + P-level) [ours]} & \text{CLAP} & LA & - & $\text{87.21}$ & 
    $\text{41.46}$ & 
    $\text{64.33}$ \\

    \midrule

    \textbf{Meta-Ensemble (Fixed Split + S-level + 2-Hidden) [ours]} & \text{CLAP} & LA & - & $\underline{\text{89.60}}_{\pm 1.43}$ & 
    $\text{43.54}_{\pm 1.55}$ & 
    $\textbf{66.49}_{\pm 0.05}$ \\
    
    \textbf{Meta-Ensemble (5-Fold + S-level + 2-Hidden) [ours]} & \text{CLAP} & LA & - & $\text{86.16}_{\pm 0.99}$ & 
    $\text{45.10}_{\pm 0.58}$ & 
    ${\text{65.63}}_{\pm 0.23}$ \\




    \bottomrule
    \end{tabular}}

\end{table*}


\subsection{Training Procedure}
The training pipeline consists of two sequential stages: base models training and meta-model training.

\subsubsection{Base Model Training}
We train five BTS-based base models under each combination of data split strategy (fixed split or 5-fold cross-validation) and partitioning level (patient-level or sample-level). Each model outputs a 4-dimensional logit vector corresponding to the four ICBHI benchmark label classes. The diversity of these outputs is crucial for downstream meta-model effectiveness.

\subsubsection{Meta-Model Training}
Once the base models are trained, we freeze them and extract their logits on the meta-model training split. The base model outputs are concatenated into a 20-dimensional vector (five models $\times$ four classes) for each sample. The meta-model takes this vector as input and produces the final respiratory class prediction.

We explore four meta-model architectures:
\begin{itemize}
    \item 1-Hidden Layer: A single fully connected layer with 512 units, followed by ReLU and a classifier.
    \item 2-Hidden Layers: Two fully connected layers with 512 units each and ReLU activations before classification.
    \item BTS-Based Meta-Model (BTS-Meta): A BTS model applied to the meta-model training set, using CLAP~\cite{wu2023large} and metadata embeddings as input.
    \item BTS + Linear Fusion (BTS-Linear): Combines BTS embeddings (1024-dim) with a linear projection of the 20-dim base logits into a 512-dim, concatenated into a 1536-dimensional feature vector for classification.
\end{itemize}

Again, all meta-models are trained using the 20\% meta-model training set, ensuring no overlap at all with data seen by the base models. This mitigates overfitting and provides a realistic evaluation of generalization to unseen data. Note that the meta-model training sets derived from both fixed split and cross-validation are the same.



\section{Experiments}
\subsection{ICBHI Dataset Comparisons}
Table~\ref{table1} presents a comprehensive comparison of prior state-of-the-art methods evaluated on the ICBHI dataset, following the official 60--40\% train-test split. 
Recent works leveraging large-scale pertaining (e.g., CLAP~\cite{wu2023large, kim24f_interspeech}, AST~\cite{gong21b_interspeech, bae23b_interspeech}, M2D-X~\cite{niizumi2024masked}) demonstrated incremental improvements over earlier CNN and ResNet-based approaches~\cite{yang2020adventitious, ma2020lungrn+, chang22h_interspeech}.
The BTS~\cite{kim24f_interspeech} achieved a strong result, with BTS++ indicating the prior best score at 65.69\%~\cite{toikkanen25_interspeech}.

Our proposed Meta-Ensemble (Fixed Split + S-level + 2-Hidden) achieved an ICBHI score of \textbf{66.49}\%, outperforming both previous methods and the best-performing Mean-ensemble variant (64.33\%) 
This shows that a learned aggregation strategy combined with diverse data splits provided superior generalization.
Interestingly, the Meta-Ensemble (5-Fold + S-level + 2-Hidden) also performed competitively, reaching 65.63\%. Although each base model was trained on a smaller portion of the data compared to the fixed split, the increased variation introduced by cross-validation compensated for this reduction. 
These results demonstrate the effectiveness of meta-ensemble learning with diverse data split strategies in advancing RSC, yielding an absolute improvement of 2.96\% points over the baseline BTS model and a further 0.82\% points gain compared to the previous state-of-the-art ensemble method~\cite{toikkanen25_interspeech}.


\begin{table}[!t]
\centering
\caption{Performance of meta-ensemble results from fixed split and 5-fold cross-validation. The \textbf{Best} and {\underline{second best}} results.}
\addtolength{\tabcolsep}{0pt}
\resizebox{\columnwidth}{!}{%
\begin{tabular}{llcc|cc}
\toprule
\multirow{2}{*}{\textbf{Split}} & \multirow{2}{*}{\textbf{Model}} & \multicolumn{2}{c|}{\textbf{Fixed Split}} & \multicolumn{2}{c}{\textbf{5-Fold}} \\
&  & Score (\%) & RRC (\%) & Score (\%) & RRC (\%) \\
\midrule
\multirow{6}{*}{\textbf{P-level}} 
& Base models (mean)     & 63.19 $\pm$ 0.28 & --      & 61.97 $\pm$ 0.84 & --      \\
& 1-Hidden               & 63.67 $\pm$ 0.56 & 0.76  & 63.64 $\pm$ 0.28 & 2.69  \\
& 2-Hidden               & 63.67 $\pm$ 0.20 & 0.76  & 63.79 $\pm$ 0.23 & \underline{2.94}  \\
& BTS-Meta            & 59.36 $\pm$ 2.63 & -6.06 & 60.83 $\pm$ 1.51 & -1.84 \\
& BTS-Linear    & 63.58 $\pm$ 0.34 & 0.62  & 63.29 $\pm$ 0.30 & 2.13  \\
\midrule
\multirow{6}{*}{\textbf{S-level}} 
& Base models (mean)     & 64.74 $\pm$ 0.38 & --      & 63.68 $\pm$ 0.44 & --      \\
& 1-Hidden               & 66.21 $\pm$ 0.32 & 2.27  & 65.43 $\pm$ 0.14 & 2.75  \\
& 2-Hidden               & \textbf{66.49} $\pm$ 0.05 & 2.70  & \underline{65.63} $\pm$ 0.23 & \textbf{3.06}  \\
& BTS-Meta            & 61.09 $\pm$ 2.77 & -5.64 & 62.01 $\pm$ 2.52 & -2.62 \\
& BTS-Linear    & 65.54 $\pm$ 0.42 & 1.24  & 65.35 $\pm$ 0.35 & 2.62  \\
\bottomrule
\end{tabular}%
\label{table2}
}

\end{table}
\begin{table}[!t]
\centering
\caption{Comparison of base models, 2-Hidden meta-ensemble, and mean-ensemble under fixed split and 5-fold cross-validation.}
\addtolength{\tabcolsep}{0pt}
\resizebox{\columnwidth}{!}{%
\begin{tabular}{llcc|cc}
\toprule
\textbf{Split} & \textbf{Model} & \multicolumn{2}{c|}{\textbf{Fixed Split}} & \multicolumn{2}{c}{\textbf{5-Fold}} \\
& & Score (\%) & RRC (\%) & Score (\%) & RRC (\%) \\
\midrule
\multirow{3}{*}{\textbf{P-level}} 
& Base model (mean) & 63.19 $\pm$ 0.28 & --    & 61.97 $\pm$ 0.84 & --    \\
& Meta-model 2-Hidden          & 63.67 $\pm$ 0.20 & 0.76  & 63.79 $\pm$ 0.23 & 2.94  \\
& Mean-ensemble     & 65.12             & 3.05  & 64.33            & 3.82  \\
\midrule
\multirow{3}{*}{\textbf{S-level}} 
& Base model (mean) & 64.74 $\pm$ 0.38 & --    & 63.68 $\pm$ 0.44 & --    \\
& Meta-model 2-Hidden          & 66.49 $\pm$ 0.05 & 2.70  & 65.43 $\pm$ 0.14 & 2.75  \\
& Mean-ensemble     & 66.34             & 2.48  & 64.03            & 0.54  \\
\bottomrule
\label{table3}
\end{tabular}}
\end{table}
\begin{table*}[!t]
\centering
\caption{Comparative studies on the ICBHI~\cite{rocha2018alpha} (in-distribution), SPRSound~\cite{zhang2022sprsound} (out-of-distribution), and SNUBH~\cite{kim2025adaptive} (in-house clinical) datasets. \textbf{Best} and \underline{second-best} results.}
\scriptsize
\resizebox{1.0\linewidth}{!}{%
\begin{tabular}{llllllllll}
\toprule
\multirow{2}{*}{Methods} & \multicolumn{3}{c}{ICBHI (In-Distribution)} & \multicolumn{3}{c}{SPRSound (Out-of-Distribution)} & \multicolumn{3}{c}{SNUBH (in-house clinical)} \\
\cmidrule(lr){2-4}\cmidrule(lr){5-7}\cmidrule(lr){8-10}
& $S_p$ (\%) & $S_e$ (\%) & \textbf{Score} (\%) & $S_p$ (\%) & $S_e$ (\%) & \textbf{Score} (\%) & $S_p$ (\%) & $S_e$ (\%) & \textbf{Score} (\%) \\
\midrule
Bae et al. \cite{bae23b_interspeech} (Fine-tuning)               & 77.14 & 41.97 & 59.55 & 69.62 & 32.65 & 51.13 & 69.47 & 67.50 & 68.49 \\
Bae et al. \cite{bae23b_interspeech} (Patch-Mix CL)               & 81.66 & 43.07 & 62.37 & 62.69 & 39.33 & 51.01 & 70.78 & 70.00 & 70.39 \\
Kim et al. \cite{kim2024stethoscope} (SG-SCL)               & 79.87 & 43.55 & 61.71 & 81.06 & 22.62 & 51.84 & 70.22 & 74.50 & 72.36 \\
Kim et al. \cite{kim24f_interspeech} (Audio-CLAP)              & 80.85 & 44.67 & 62.56 & 70.67 & \underline{\text{41.90}} & 56.29 & 70.58 & \textbf{80.00} & 75.29 \\
Kim et al. \cite{kim24f_interspeech} (BTS)        & 81.40 & \underline{\text{45.67}} & \text{63.54} & 67.50 & 39.33 & 53.42 & 77.38 & 76.14 & 76.76 \\

\textbf{Meta-Ensemble (Fixed Split + S-level)} & $\textbf{{89.60}}_{\pm 1.43}$ & $\text{43.54}_{\pm 1.55}$ & $\textbf{66.49}_{\pm 0.05}$ & $\underline{\text{{84.76}}}_{\pm 1.45}$ & $\text{32.39}_{\pm 2.02}$ & $\text{58.57}_{\pm 0.89}$ & $\text{76.40}_{\pm 3.33}$ & $\text{74.00}_{\pm 1.37}$ & $\text{75.20}_{\pm 2.26}$ \\

\textbf{Meta-Ensemble (Fixed Split + P-level)} & $\text{{83.86}}_{\pm 2.18}$ & $\text{43.49}_{\pm 1.99}$ & $\text{63.67}_{\pm 0.20}$ & $\textbf{{85.50}}_{\pm 2.36}$ & $\text{38.10}_{\pm 5.69}$ & $\underline{\text{61.80}}_{\pm 1.72}$ & ${\text{80.24}}_{\pm 4.59}$ & $\underline{\text{77.50}}_{\pm 2.50}$ & $\underline{\text{78.82}}_{\pm 1.38}$ \\

\textbf{Meta-Ensemble (5-Fold + S-level)} & $\underline{\text{86.16}}_{\pm 0.99}$ & $\text{45.10}_{\pm 0.58}$ & $\underline{\text{65.63}}_{\pm 0.23}$ & $\text{{78.50}}_{\pm 5.58}$ & $\text{39.37}_{\pm 3.88}$ & $\text{58.93}_{\pm 1.65}$ & $\underline{\text{83.55}}_{\pm 2.19}$ & $\text{73.00}_{\pm 3.26}$ & $\text{78.28}_{\pm 1.40}$ \\ 

\textbf{Meta-Ensemble (5-Fold + P-level)} & $\text{{80.65}}_{\pm 4.02}$ & $\textbf{46.93}_{\pm 3.75}$ & $\text{63.79}_{\pm 0.23}$ & $\text{{79.14}}_{\pm 3.71}$ & $\textbf{\text{44.56}}_{\pm 2.87}$ & $\textbf{61.85}_{\pm 0.47}$ & $\textbf{88.78}_{\pm 3.55}$ & $\text{69.50}_{\pm 4.47}$ & $\textbf{79.14}_{\pm 1.33}$ \\ 

\bottomrule
\end{tabular}
}
\label{table4}
\end{table*}

\subsection{Effect of Split Strategies on Meta-Ensemble Performance}
Table~\ref{table2} summarizes the performance of the proposed meta-ensemble approach under two data splitting strategies. Here, \textit{Base models (mean)} denotes the average performance of the five independently trained base models for each split configuration. 

At the \textbf{P-level}, the best performance under fixed split was achieved by the 1-Hidden and 2-Hidden meta-models (63.67\%), showing only marginal gains over the base models. 
When using 5-fold cross-validation, the 2-Hidden meta-model achieved the highest score (63.79\%, RRC +2.94\%), while BTS-Meta again yielded negative gains (RRC -1.84\%). 


At the \textbf{S-level}, the improvements were more dramatic. Under fixed split, the 2-Hidden meta-model achieved the best performance (66.49\%, RRC +2.70\%), followed closely by the 1-Hidden model (RRC +2.27\%). 
With 5-fold cross-validation, the 1-Hidden meta-model obtained the best result (65.63\%, RRC +3.06\%), followed by the 2-Hidden model (RRC +2.75\%). Again, BTS-Meta failed to improve over base ensemble. 

Overall, the base model mean performance was consistently higher under fixed split than 5-fold cross-validation, and P-level generally outperformed S-level. This indicates that patient overlap across splits strongly influences classification accuracy in the ICBHI dataset. While absolute ICBHI scores appeared superior under fixed split, RRC analysis revealed that 5-fold cross-validation yielded stronger relative improvements over the base models, reflecting its role in enhancing data utilization and base model diversity.
Moreover, the consistent superiority of 1- and 2-Hidden meta-models suggests that lightweight non-linear transformations are more effective for combining diverse logits than direct concatenation or BTS-based fusion. These results highlight that S-level splits may exaggerate performance due to patient overlap in the in-distribution dataset, whereas P-level results provide a stricter and more clinically relevant assessment of generalization, like in OOD settings.

\subsection{Comparison with Mean-Ensemble}
To ablate the effectiveness of the proposed meta-ensemble framework, we compare it with the mean-ensemble, where the logits are combined by averaging. Several important patterns have emerged from Table~\ref{table3}. First, while the mean-ensemble consistently improved over the base models, the 2-Hidden meta-ensemble generally provided additional gains in RRC. Notably, the 5-fold setting often amplified these improvements, particularly at the P-level, where the 2-Hidden meta-ensemble and mean-ensemble achieved RRCs of +2.94\% and +3.82\%, respectively, outperforming their fixed split counterparts. Although in the S-level condition, the mean-ensemble under 5-fold showed a relatively small RRC (+0.54\%), the 2-Hidden meta-ensemble still maintained a clear gain (+2.75\%), emphasizing the advantage of learned aggregation under cross-validation. These findings suggest that 5-fold cross-validation enhances the effectiveness and stability of both mean-ensemble approaches by increasing data diversity and reducing variance across splits. 

Although the mean-ensemble showed stronger RRC improvements at the P-level under both fixed split (+3.05\%) and 5-fold (+3.82\%) compared to the meta-ensemble (+0.76\% and +2.94\%, respectively), the meta-ensemble consistently outperformed the mean-ensemble at the S-level. This indicates that the meta-ensemble particularly excels when handling fine-grained sample-level variation, while the mean-ensemble still retains competitive strength at the patient-level. Overall, the benefits of meta-ensembles became most pronounced when combined with the diversity introduced by 5-fold cross-validation, especially in the S-level setting. 


\subsection{Evaluations on Out-of-Distribution Datasets}
To demonstrate the generalizability of our meta-ensemble approach, we evaluated on two OOD datasets, SPRSound~\cite{zhang2022sprsound} and SNUBH~\cite{kim2025adaptive}, which differ in demographics, recording protocols, and labeling granularity, providing a strong test of cross-dataset generalization.



Table~\ref{table4} illustrates the key strength of our approach in OOD settings. On the in-distribution ICBHI dataset, the best and second-best results were obtained under the S-level configuration (66.49\% and 65.63\%), demonstrating that patient overlap inflates in-distribution scores without reflecting true generalization.
By contrast, the trend reversed in OOD evaluations: P-level configuration consistently yielded the strongest generalization. On SPRSound, the P-level meta-ensembles achieved the best and second-best results, surpassing all prior methods.
While the difference between fixed split and 5-fold cross-validation was relatively small, the gap between P-level and S-level was substantial, highlighting that enforcing patient non-overlap is the critical factor for robust OOD generalization.

On the SNUBH clinical dataset, the contrast was even clearer. The baseline BTS scored 76.76\%, but the Meta-Ensemble with fixed + S-level dropped to 75.20\%, showing that sample-level configurations 
degrade clinical performance despite their apparent advantage in the in-distribution setting. In contrast, P-level settings consistently outperformed the baseline: Fixed + P-level reached 78.82\% (second-best), and 5-Fold + P-level achieved the best score of 79.14\%. These findings suggest that while S-level splits may exaggerate in-distribution performance, only P-level configurations consistently provide reliable generalization in OOD environments. 


\section{Conclusion}
We presented a meta-ensemble learning approach for RSC that leverages data diversity through different split strategies. Our results showed that combining predictions from varied base models using a learned meta-model improved generalization, achieving new state-of-the-art performance on the ICBHI dataset and effectively transferring to the unseen OOD datasets.

\bibliographystyle{IEEEtran}
\small
\bibliography{refs}

\end{document}